\documentclass[11pt,letterpaper]{article}
\usepackage{amsmath,amssymb}
\usepackage{amsthm} 
\usepackage{Alex-setup,times}
\FULLPAGE

\usepackage{natbib}
\usepackage{natbibspacing}
\usepackage{booktabs}
\usepackage{cases}
\usepackage{dsfont}
\usepackage{epstopdf}
\usepackage{algorithm}
\usepackage{algpseudocode}

\newcommand{\polylog}{\mathrm{polylog}}

\newlength{\minipagewidth}
\newcommand{\bookbox}[1]{\small
\par\medskip\noindent
\framebox[\columnwidth]{
\begin{minipage}{\minipagewidth} {#1} \end{minipage} } \par\medskip }

\def\ds1{\mathds{1}}
\renewcommand{\phi}{\varphi}
\renewcommand{\epsilon}{\varepsilon}
\newcommand{\Ber}{\mathtt{Bernoulli}}

\newcommand{\alg}{{\texttt{SAO}}}
\newcommand{\tG}{\tilde{G}_{i,t}}

\newcommand{\G}{G_{i,t}}
\newcommand{\tH}{\tilde{H}_{i,t}}
\newcommand{\hH}{\widehat{H}_{i,t}}
\newcommand{\oH}{H_{i,t}}

\newcommand{\oR}{\overline{R}}
\newcommand{\E}{\mathbb{E}}

\newcommand{\cS}{\mathcal{S}}

\newcommand{\cF}{\mathcal{F}}

\renewcommand{\P}{\mathbb{P}}

\newcommand{\argmax}{\mathop{\mathrm{argmax}}}

\newcommand{\bigE}[1]{\E\left[\, #1 \,\right]} 

\newcommand{\estimate}{\widetilde}
\newcommand{\algo}{\widehat}

\newcommand{\eps}{\epsilon}

\newcommand{\UCB}{\ensuremath{\mathtt{UCB1}}}
\newcommand{\EXP}{\ensuremath{\mathtt{Exp3}}}
\newcommand{\EXPp}{\ensuremath{\mathtt{Exp3.P}}}

\begin{document}

\title{The best of both worlds: stochastic and adversarial bandits}

\date{February 2012}

\author{
S{\'e}bastien Bubeck
\footnote{Department of Operations Research and Financial Engineering, Princeton University,
Princeton, NJ, USA. Email: {sbubeck@princeton.edu}.}
\and
Aleksandrs Slivkins%
\footnote{Microsoft Research, Mountain View, USA.
Email: slivkins@microsoft.com.}
}

\maketitle

\begin{abstract}
We present a bandit algorithm, {\alg} (Stochastic and Adversarial Optimal), whose regret is, essentially, optimal both for adversarial rewards and for
stochastic rewards. Specifically, {\alg} combines the $O(\sqrt{n})$ worst-case regret of Exp3 \citep{bandits-exp3} for adversarial rewards and the (poly)logarithmic regret of UCB1~\citep{bandits-ucb1} for stochastic rewards.  Adversarial rewards and stochastic rewards are the two main settings in the literature on (non-Bayesian) multi-armed bandits. Prior work on multi-armed bandits treats them separately, and does not attempt to jointly optimize for both. Our result falls into a general theme of achieving  good worst-case performance while also taking advantage of ``nice'' problem instances, an important issue in the design of algorithms with partially known inputs.
\end{abstract}


\section{Introduction}

Multi-armed bandits (henceforth, MAB) is a simple model for sequential decision making under uncertainty that captures the crucial tradeoff between \emph{exploration} (acquiring new information) and \emph{exploitation} (optimizing based on the information that is currently available). Introduced in early 1950-ies \citep{Robbins1952}, it has been studied intensively since then in Operations Research, Electrical Engineering, Economics, and Computer Science.

\begin{figure}[t]
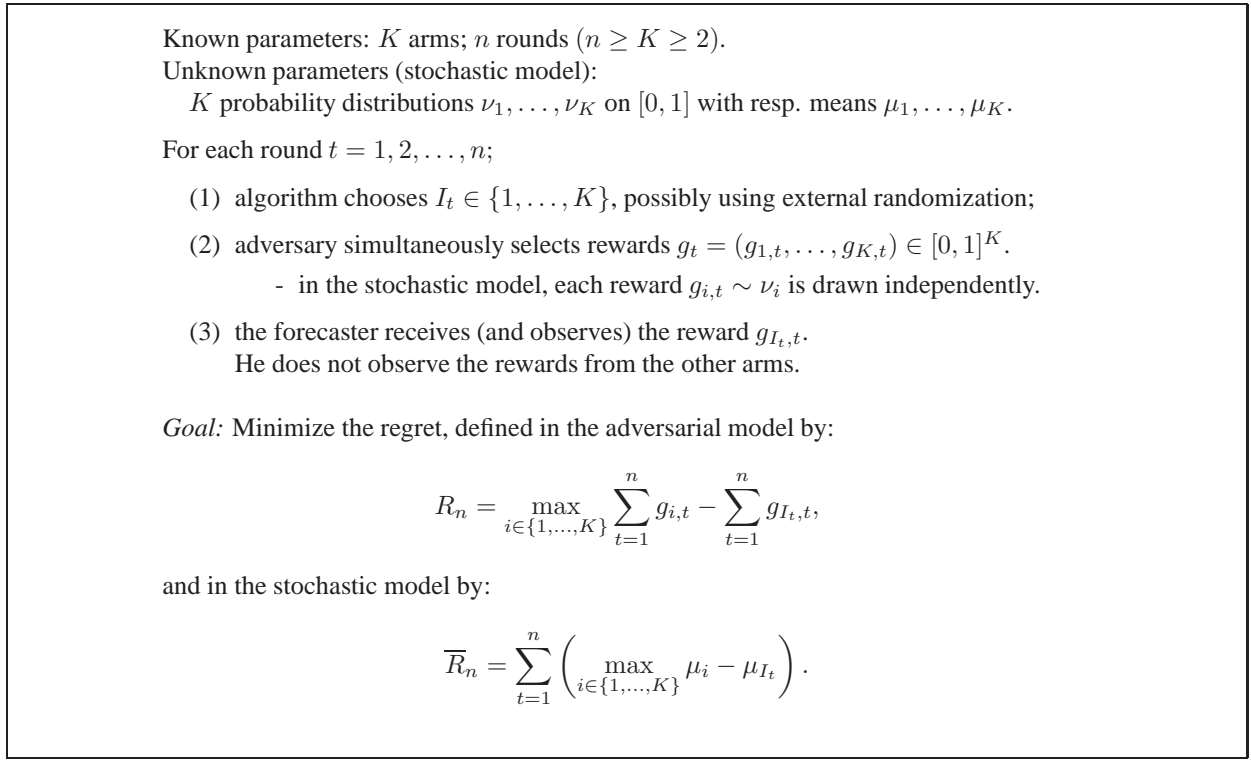

\bookbox{
Known parameters: $K$ arms; $n$ rounds $(n \geq K \geq 2)$.\\
Unknown parameters (stochastic model):

~~~~$K$ probability distributions $\nu_1,\ldots,\nu_K$ on $[0,1]$
with resp. means $\mu_1, \hdots, \mu_K$.

\medskip\noindent
For each round $t=1,2,\ldots,n;$
\begin{itemize}
\item[(1)]
algorithm chooses $I_t \in \{1,\ldots,K \}$, possibly using external randomization;



\item[(2)] adversary simultaneously selects rewards
            $g_t=(g_{1,t},\hdots,g_{K,t}) \in [0,1]^K$.
\begin{OneLiners}
\item[-] in the stochastic model, each reward $g_{i,t} \sim \nu_i$ is drawn independently.
\end{OneLiners}

\item[(3)]
the forecaster receives (and observes) the reward $g_{I_t,t}$.\\ He does not observe the rewards from the other arms.
\end{itemize}

\medskip\noindent
{\em Goal:} Minimize the regret, defined in the adversarial model by:
$$R_n = \max_{i \in \{1,\hdots,K\}} \sum_{t=1}^n g_{i,t} - \sum_{t=1}^n g_{I_t,t},$$
and in the stochastic model by:
$$\oR_n = \sum_{t=1}^n  \left(\max_{i \in \{1,\hdots,K\}} \mu_i - \mu_{I_t}\right).$$
}
\caption{The MAB framework: adversarial rewards and stochastic rewards.}
\label{fig:game}
\end{figure}

The ``basic" MAB framework can be formulated as a game between the player (i.e., the algorithm) and the adversary (i.e., the environment). The player selects actions (``arms") sequentially from a fixed, finite set of possible options, and receives rewards that correspond to the selected actions. For simplicity, it is customary to assume that the rewards are bounded in $[0,1]$. In the adversarial model one makes no other restrictions on the sequence of rewards, while in the stochastic model we assume that the rewards of a given arm is an i.i.d sequence of random variables. The performance criterion is the so-called
regret, which compares the rewards received by the player to the rewards accumulated by a hypothetical benchmark algorithm. A typical, standard benchmark is the best single arm. See Figure \ref{fig:game} for a precise description of this framework.

Adversarial rewards and stochastic rewards are the two main reward models in the MAB literature. Both are now very well understood, in particular thanks to the seminal papers \citep{Lai-Robbins-85,bandits-ucb1,bandits-exp3}.
In particular, the $\EXP$ algorithm from \citep{bandits-exp3} attains a regret growing as $O(\sqrt{n})$ in the adversarial model, where $n$ is the number of rounds, and $\UCB$ algorithm from \citep{bandits-ucb1} attains $O(\log n)$ in the stochastic model. Both results are essentially optimal. It is worth noting that $\UCB$ and $\EXP$ have influenced, and to some extent inspired, a number of follow-up papers on richer MAB settings.

However, it is easy to see that $\UCB$ incurs a trivial $\Omega(n)$ regret in the adversarial model, whereas $\EXP$ has $\Omega(\sqrt{n})$ regret even in the stochastic model.%
\footnote{This is clearly true for the original version of $\EXP$ with a mixing parameter. However, this mixing is unnecessary against oblivious adversaries \citep{Sto05}. The regret of the resulting algorithm
in the stochastic model is unknown.} This raises a natural question that we aim to resolve in this paper: \emph{can we achieve the best of both worlds?} Is there a bandit algorithm which matches the performance of $\EXP$ in the adversarial model, and attains the performance of $\UCB$ if the rewards are in fact stochastic? A more specific (and slightly milder) formulation is as follows:
\begin{quote}
Is there a bandit algorithm that has $\tilde{O}(\sqrt{n})$ regret in the adversarial model and $\polylog(n)$ regret in the stochastic model?
\end{quote}
We are not aware of any prior work on this question. Intuitively, we introduce a new tradeoff: a bandit algorithm has to balance between \emph{attacking} the weak adversary (stochastic rewards) and \emph{defending} itself from a more devious adversary that targets algorithm's weaknesses, such as being too aggressive if the reward sequence is seemingly stochastic. In particular, while the basic exploration-exploitation tradeoff induces $O(\log n)$ regret in the stochastic model, and $O(\sqrt{n})$ regret in the adversarial model, it is not clear a priori what are the optimal regret guarantees for this new \emph{attack-defense} tradeoff.

We answer the above question affirmatively, with a new algorithm called {\alg} (Stochastic and Adversarial Optimal). To formulate our result, we need to introduce some notation. In the stochastic model, let $\mu_i$ be the expected single-round reward from arm $i$. A crucial parameter is the \emph{minimal gap}:
    $\Delta = \min_{i:\; \mu_i<\mu^*} \mu^* - \mu_i$,
where
    $\mu^* = \max_i \mu_i$.
With this notation, $\UCB$ attains regret $O(\tfrac{K}{\Delta} \log n)$ in the stochastic model, where $K$ is the number of arms.
We are looking for the following:  regret
    $\E[R_n] = \tilde{O}(\sqrt{Kn})$
in the adversarial model and regret
    $\E[\oR_n] = \tilde{O}(\tfrac{K}{\Delta})$
in the stochastic model, where $\tilde{O}(\cdot)$ hides $\polylog(n)$ factors.
Our main result is as follows.

\begin{theorem}\label{thm:intro}
There exists an algorithm {\alg} for the MAB problem (Algorithm~\ref{alg:soa} on page~\pageref{alg:soa}) such that:
\begin{OneLiners}
\item[(a)] in the adversarial model, {\alg} achieves regret
        $\E[R_n] \leq O(\sqrt{nK}\, \log^{3/2}(n)\, \log K)$.
\item[(b)] in the stochastic model, {\alg} achieves regret
        $\E[\oR_n] \leq O(\tfrac{K}{\Delta}\, \log^2(n) \log K )$.
\end{OneLiners}
\end{theorem}

\noindent Moreover, with very little extra work we can obtain the corresponding high-probability versions (see Theorem \ref{th:1} for a precise statement).

It is easier, and more instructive, to explain the main ideas on the special case of two arms and oblivious adversary.\footnote{An oblivious adversary fixes the rewards $g_{i,t}$ for all round $t$ without observing the algorithm's choices.} This special case (with a simplified algorithm) is presented in Section~\ref{sec:2-arms}. The general case is then fleshed out in Section~\ref{sec:K-arms}.



\xhdr{Discussion.}
The question raised in this paper touches upon an important theme in Machine Learning, and more generally in the design of algorithms with partially known inputs: how to achieve a good worst-case performance \emph{and} also take advantage of ``nice'' problem instances. In the context of MAB it is natural to focus on the distinction between stochastic and adversarial rewards, especially given the prominence of the two models in the MAB literature. Then our ``best-of-both-worlds'' question is the first-order specific question that one needs to resolve. Also, we provide the first analysis of the same MAB algorithm under both adversarial and stochastic rewards.

Once the ``best-of-both-worlds'' question is settled,
several follow-up questions emerge. Most immediately, it is not clear whether the polylog factors can be improved to match the optimal guarantees for each respective model; a lower bound would indicate that the ``attack-defence'' tradeoff is fundamentally different from the familiar explore-exploit tradeoffs.
A natural direction for further work is rewards that are adversarial on a few short time intervals, but stochastic most of the time. Moreover, it is desirable to adapt not only to the binary distinction between the stochastic and adversarial rewards, but also to some form of continuous tradeoff between the two reward models.

Finally, we acknowledge that our solution is no more (and no less) than a theoretical proof of concept. More work, theoretical and experimental, and perhaps new ideas or even new algorithms, are needed for a practical solution. In particular, a practical algorithm  should probably go beyond what we accomplish in this paper, along the lines of the two possible extensions mentioned above.

\xhdr{Related work.}
The general theme of combining worst-case and optimistic performance bounds have received considerable attention in prior work on online learning.  A natural incarnation of this theme in the context of MAB concerns proving upper bounds on regret that can be written in terms of some complexity measure of the rewards, and match the optimal worst-case bounds. To this end, a version of $\EXP$ achieves regret $\tilde{O}(\sqrt{K G^*_n})$, where  $G^*_n\leq n$ is the maximal cumulative reward of a single arm, and the corresponding high probability result
was recently proved in \citet{Bubeck-colt09}. In \citet{Hazan-soda09}, the authors obtain regret $\tilde{O}(\sqrt{K V^*_n})$, where  $V^*_n\leq n$ is the maximal ``temporal variation" of the rewards.\footnote{The result in \citet{Hazan-soda09} does not shed light on the question in the present paper, because the ``temporal variation'' concerns actual rewards rather than expected rewards. In particular, temporal variation is minimal when the actual reward of each arm is constant over time, and (essentially) maximal in the stochastic model with 0-1 rewards.} Similar results have been obtained for the full-feeback (``experts") version in \citet{Stoltz-colt05,AbernethyHR-colt08}. Also, the regret bound for $\UCB$ depends on the gap $\Delta$, and matches the optimal worst-case bound for the stochastic model (up to logarithmic factors).  Moreover, adaptivity to ``nice" problem instances is a crucial theme in the work on bandits in metric spaces \citep{LipschitzMAB-stoc08,xbandits-nips08,contextualMAB-colt11}, an MAB setting in which some information on similarity between arms is a priori available to an algorithm.

The distinction between $\polylog(n)$ and $\Omega(\sqrt{n})$ regret has been crucial in other MAB settings:
bandits with linear rewards \citep{DaniHK-colt08},
bandits in metric spaces \citep{DichotomyMAB-soda10},
and an extension of MAB to auctions \citep{MechMAB-ec09,DevanurK08,Transform-ec10}. Interestingly, here we have four different MAB settings (including the one in this paper) in which this distinction occurs for four different reasons, with no apparent connections.

A proper survey of the literature on multi-armed bandits is beyond the scope of this paper; a reader is encouraged to refer to \cite{CesaBL-book} for background. An important high-level distinction is between Bayesian and non-Bayesian MAB formulations. Both have a rich literature; this paper focuses on the latter. The ``basic" MAB version defined in this paper has been extended in various papers to include additional information and/or assumptions about rewards.

Most relevant to this paper are algorithms $\UCB$ \citep{bandits-ucb1} and $\EXP$ \citep{bandits-exp3}. $\UCB$ has a slightly more refined regret bound than the one that we cited earlier:
    $\oR_n = O(\sum_{i:\, \mu_i<\mu^*} \frac{\log n}{\mu^*-\mu_i} )$
with high probability. A matching lower bound (up to the considerations of the variance and constant factors) is proved in
\citet{Lai-Robbins-85}. Several recent papers \citep{AO10,HT10,AMS09,Bubeck-colt09,Maillard-colt11,Garivier-colt11,PR11} improve over $\UCB$, obtaining algorithms with regret bounds that are even closer to the lower bound.

The regret bound for $\EXP$ is
    $\E[R_n] = O(\sqrt{nK \log K})  $,
and a version of $\EXP$ achieves this with high probability \citep{bandits-exp3}. There is a nearly matching lower bound of $\Omega(\sqrt{Kn})$.
Recently \citet{Bubeck-colt09} have shaved off the $\log K$ factor, achieving an algorithm with regret
    $O(\sqrt{Kn})$
in the adversarial model against an oblivious adversary.



\xhdr{High-level ideas.}
For clarity, let us consider the simplified algorithm for the special case of two arms and oblivious adversary. The algorithm starts with the assumption that the stochastic model is true, and then proceeds in three phases: ``exploration", ``exploitation", and the ``adversarial phase".
In the exploration phase, we alternate the two arms until one of them (say, arm 1) appears significantly better than the other. When and if that happens, we move to the exploitation phase
where we focus on arm 1, but re-sample arm $2$ with small probability.
After each round we check several consistency conditions which should hold with high
probability if the rewards are stochastic. When and if one of these conditions fails, we declare that we are not in the case of stochastic rewards, and switch to running a
bandit algorithm for the adversarial model (a version of $\EXP$).

Here we have an incarnation of the ``attack-defense'' tradeoff mentioned earlier in this section: the consistency conditions should be (a) strong enough to justify using the stochastic model as an operating assumption while the conditions hold, and (b) weak enough so that we can check them despite the low sampling probability of arm $2$. The fact that (a) and (b) are not mutually exclusive is surprising and unexpected.

More precisely, the consistency conditions should be strong enough to insure us from losing too much in the first two phases even if we are in the adversarial model. We use a specific re-sampling schedule for arm $1$ which is rare enough so that we do not accumulate much regret if this is indeed a bad arm, and yet sufficient to check the consistency conditions.

To extend to the $K$-arm case, we ``interleave" exploration and exploitation, ``deactivating" arms one by one as they turn out to be suboptimal. The sampling probability of a given arm increases while the arm stays active, and then decreases after it is deactivated, with a smooth transition between the two phases. This complicated behavior (and the fact that we handle general adversaries) in turn necessitate a more delicate analysis.

\section{Preliminaries}
\label{sec:prelims}


We consider randomized algorithms, in the sense that $I_t$ (the arm chosen at time $t$) is drawn from a probability distribution $p_t$ on $\{1,\hdots,K\}$. We denote by $p_{i,t}$ the probability that
$I_t = i$. For brevity, let
    $I_{i,t} = \ds1_{\{I_t=i\}}$.
Given such a randomized algorithm, it is a well-known trick to use
    $\tilde{g}_{i,t} = \frac{g_{i,t}\, I_{i,t} }{p_{i,t}}$
as an unbiased estimate of the reward $g_{i,t}$.
Now for arm $i$ and time $t$ we introduce:
\begin{itemize}
\item $G_{i,t} = \sum_{s=1}^t g_{i,s}$
    ~~~~~~~(\emph{fixed-arm cumulative reward}
     from arm $i$ up to time $t$),
\item $\estimate{G}_{i,t} = \sum_{s=1}^t \tilde{g}_{i,s}$
    ~~~~~~~(\emph{estimated cumulative reward} from arm $i$ up to time $t$),
\item $\algo{G}_{i,t} = \sum_{s=1}^t g_{i,s}\, I_{i,s}$
    ~~~(\emph{algorithm's cumulative reward} from arm $i$ up to time $t$),
\item $T_i(t) = \sum_{s=1}^{t} I_{i,s}$
    ~~~~~~(the \emph{sampling time} of arm $i$ up to time $t$).
\item The corresponding averages:
    $H_{i,t} = \tfrac{1}{t}\, G_{i,t}$,
    $\estimate{H}_{i,t} = \tfrac{1}{t}\, \estimate{G}_{i,t}$, and
    $\algo{H}_{i,t} = \algo{G}_{i,t}/ T_i(t)$.
\end{itemize}

$G_{i,t}$ is the cumulative reward of a ``fixed-arm algorithm" that always plays arm $i$. Recall that our benchmarks are
    $\max_i\, G_{i,t}$ for the adversarial model, and
    $\max_i\, \E[G_{i,t}]$
for the stochastic model.

Note that $\estimate{H}_{i,t}$, $\algo{H}_{i,t}$ (and $\estimate{G}_{i,t}$, $\algo{G}_{i,t}$) are observed by an algorithm whereas $H_{i,t}$ (and $G_{i,t}$) is not. Informally, $\estimate{H}_{i,t}$ and $\algo{H}_{i,t}$ are estimates for the expected reward $\mu_i$ in the stochastic model, and $\estimate{H}_{i,t}$ is an estimate for the benchmark reward $H_{i,t}$ in the adversarial model.

In the stochastic model we define the \emph{gap} of arm $i$ as
    $\Delta_i = (\max_{1 \leq j \leq K} \mu_j) - \mu_i$,
and the \emph{minimal gap}
$    \Delta = \min_{i:\, \Delta_i>0} \Delta_i$.

\OMIT{
cumulative gain $G_{i,t} = \sum_{s=1}^t g_{i,s}$,
estimated cumulative gain $\estimate{G}_{i,t} = \sum_{s=1}^t \estimate{g}_{i,s}$,
and the algorithm's cumulative gain: $\algo{G}_{i,t} = \sum_{s=1}^t g_{i,s} \ds1_{I_s=i}$. Moreover, the sampling time of arm $i$ at time $t$ is defined as
    $T_i(t) = \sum_{s=1}^{t} \ds1_{I_s=i}$.
} 

Following the literature, we measure algorithm's performance in terms of \emph{regret} $R_n$ and $\oR_n$ as defined in Figure~\ref{fig:game}. The two notions of regret are somewhat different, in particular the ``stochastic regret'' $\oR_n$ is \emph{not} exactly equal to the expected ``adversarial regret'' $R_n$. However, in the stochastic model they are approximately equal:%
\footnote{This fact is well-known and easy to prove, e.g. see Proposition~34 in~\citet{Bubeck-colt09}.}
$ \E[\oR_n] \leq \E[R_n] \leq \E[\oR_n] + \sqrt{\tfrac12\;n \log K}$.

\section{A simplified SAO algorithm for $K=2$ arms}
\label{sec:2-arms}

\newcommand{\chernoff}{C_\mathtt{crn}}
\newcommand{\half}{\mathtt{hf}}
\newcommand{\TimeX}{\tau_*}
\newcommand{\INT}{\mathtt{INT}}

\newcommand{\condSpace}{post-exploration probability space}

\newcommand{\simpleAlg}{\texttt{SimpleSAO}}

We will derive a (slightly weaker version of) the main result for the special case of $K=2$ arms and oblivious adversary, using a simplified algorithm. This version contains most of the ideas from the general case, but can be presented in a more lucid fashion.

We are looking for the ``best-of-both-worlds" feature: $\tilde{O}(\sqrt{n})$ regret in the adversarial model, and
    $\tilde{O}(\tfrac{1}{\Delta})$
regret in the stochastic model, where $\Delta = |\mu_1-\mu_2|$ is the gap. Our goal in this section is to obtain this feature in the simplest way possible. In particular, we will hide the constants under the $O()$ notation, and will not attempt to optimize the $\text{polylog}(n)$ factors; also, we will assume oblivious adversary.
We will prove the following theorem:

\begin{theorem}\label{thm:2-arms}
Consider a MAB problem with two arms. There exists an algorithm such that:
\begin{OneLiners}
\item[(a)] against an oblivious adversary, its expected regret is
        $\E[R_n] \leq O(\sqrt{n}\, \log^2 n)$.
\item[(b)] in the stochastic model, its expected regret satisfies
        $\E[\oR_n] \leq O(\tfrac{1}{\Delta}\, \log^3 n  )$.
\end{OneLiners}
Both regret bounds also hold with probability at least $1-\tfrac{1}{n}$.
\end{theorem}

Note that in the stochastic model, regret trivially cannot be larger than $\Delta n$, so part (b) trivially implies regret $\E[\oR_n] \leq \tilde{O}(\sqrt{n})$.

Our analysis proceeds via high-probability arguments and directly obtains the high-probability guarantees. The high-probability arguments tend to make the analysis cleaner; we suspect it cannot be made much simpler if we only seek bounds on expected regret.

\subsection{A simplified SAO (Stochastic and Adversarial Optimal) Algorithm}
\label{subsec:simpleSAO}

The algorithm proceeds in three phases: exploration, exploitation, and adversarial phase. In the exploration phase, we alternate the two arms until one of them appears significantly better than the other. In exploitation phase, we focus on the better arm, but re-sample the other arm with small probability. We check several consistency conditions which should hold with high probability if the rewards are stochastic. When and if one of these conditions fails, we declare that we are not in the case of stochastic rewards, and switch to running a bandit algorithm for the adversarial model, namely algorithm $\EXPp$~\citep{bandits-exp3}.

The algorithm is parameterized by $\chernoff=\Theta(\log n)$ which we will chose later in Section~\ref{subsec:2arms-chernoff}. The formal description of the three phases is as follows.

\begin{description}
\item[(Exploration phase)] In each round $t$, pick an arm at random:
    $p_{1,t}=p_{2,t} = \tfrac12$.
Go to the next phase as soon as
    $t>\Omega(\chernoff^2)$
and the following condition fails:
\begin{align}\label{eq:cond-exploration}
    |\estimate{H}_{1,t}- \estimate{H}_{2,t}|< 24\,\chernoff /\sqrt{t}.
\end{align}
Let $\TimeX$ be the duration of this phase. Without loss of generality, assume
    $\estimate{H}_{1, \TimeX}> \estimate{H}_{2, \TimeX}$.
This means, informally, that arm $1$ is selected for exploitation.

\item[(Exploitation phase)]
In each round $t> \TimeX$, pick arm 2 with probability
    $p_{2,t} = \tfrac{\TimeX}{2t}$,
and arm $1$ with the remaining probability
    $p_{1,t} = 1-\tfrac{\TimeX}{2t}$.

After the round, check the following \emph{consistency conditions}:
\begin{align}
8\,\chernoff /\sqrt{\TimeX}
    \leq \estimate{H}_{1,t}- \estimate{H}_{2,t}
    \leq 40\,\chernoff /\sqrt{\TimeX}
    \label{eq:cond1} \\
\begin{cases}
|\estimate{H}_{1,t} - \algo{H}_{1,t}| \leq 6\,\chernoff /\sqrt{t} \\
|\estimate{H}_{2,t} - \algo{H}_{2,t}| \leq 6\,\chernoff /\sqrt{\TimeX} \,.
\end{cases}
    \label{eq:cond2}
\end{align}
If one of these conditions fails, go to the next phase.
\item[(Adversarial phase)] Run algorithm $\EXPp$ from~\cite{bandits-exp3}.
\end{description}

\xhdr{Discussion.} The exploration phase is simple: Condition~\eqref{eq:cond-exploration} is chosen so that once it fails then (assuming stochastic rewards) the seemingly better arm is indeed the best arm with high probability.

In the exploitation phase, we define the re-sampling schedule for arm $2$ and a collection of ``consistency conditions''. The re-sampling schedule should be sufficiently rare to avoid accumulating much regret if arm $2$ is indeed a bad arm. The consistency conditions should be sufficiently strong to justify using the stochastic model as an operating assumption while they hold. Namely, an adversary constrained by these conditions should not be able to inflict too much regret on our algorithm in the first two phases. Yet, the consistency conditions should be weak enough so that they hold with high-probability in the stochastic model, despite the low sampling probability of arm $2$.

It is essential that we use both $\algo{H}_{i,t}$ and $\estimate{H}_{i,t}$ in the consistency conditions: the interplay of these two estimators allows us to bound regret in the adversarial model. Other than that, the conditions that we use are fairly natural (the surprising part is that they work). Condition~\eqref{eq:cond1} checks whether the relation between the two arms is consistent with the outcome of the exploration phase, i.e. whether arm $1$ still seems better than arm $2$, but not \emph{too much} better. Condition~\eqref{eq:cond2} checks whether for each arm $i$, the estimate $\estimate{H}_{i,t}$ is close to the average $\algo{H}_{i,t}$. In the stochastic model, both estimate the expected gain $\mu_i$, so we expect them to be not too far apart. However, our definition of ``too far'' should be consistent with how often a given arm is sampled.

\OMIT{
\begin{align}
\estimate{H}_{1,t}- \estimate{H}_{2,t}    &\geq 8\,\chernoff /\sqrt{\TimeX}
    \label{eq:cond1}\\
\estimate{H}_{1,t}- \estimate{H}_{2,t}    &\leq 40\,\chernoff /\sqrt{\TimeX}
    \label{eq:cond1}\\
|\estimate{H}_{1,t} - \algo{H}_{1,t}| &\leq 6\,\chernoff /\sqrt{t}
    \label{eq:cond-p2-arm1} \\
|\estimate{H}_{2,t} - \algo{H}_{2,t}| &\leq 6\,\chernoff /\sqrt{\TimeX} \,.
    \label{eq:cond2}
\end{align}
} 

\subsection{Concentration inequalities}
\label{subsec:2arms-chernoff}

The ``probabilistic'' aspect of the analysis is confined to proving that several properties of estimates and sampling times hold with high probability. The rest of the analysis can proceed \emph{as if} these properties hold with probability $1$. In particular, we have made our core argument essentially deterministic, which greatly simplifies presentation.

All high-probability results are obtained using an elementary concentration inequality loosely known as \emph{Chernoff Bounds}. For the sake of simplicity, we use a slightly weaker formulation below (see Appendix~\ref{app:chernoff} for a proof), which uses just one inequality for all cases.

\begin{theorem}[Chernoff Bounds]\label{thm:our-chernoff}
Let $X_t$, $t\in [n]$ be a independent random variables such that $X_t\in [0,1]$ for each $t$. Let $X = \sum_{t=1}^n X_t$ be their sum, and let $\mu = \E[X]$. Then
\begin{align}\label{eq:our-chernoff}
\Pr\left[\, |X - \mu| > C\,\max(1,\sqrt{\mu}) \,\right] < 2\,e^{-C/3}, \quad
    \text{for any $C>1$.}
\end{align}
\end{theorem}

\OMIT{
The above formulation is slightly non-standard because of the $\max$ term in~\eqref{eq:our-chernoff} which allows for a stronger guarantee for small $\mu$. A proof can be found in Appendix~\ref{app:chernoff}.}

We will often need to apply Chernoff Bounds to sums whose summands depend on some events in the execution of the algorithms and therefore are not mutually independent. However, in all cases these issues are but a minor technical obstacle which can be side-stepped using a slightly more careful setup.%
\footnote{However, the independence issues appear prohibitive for $K>2$ arms or if we consider a non-oblivious adversary. So for the general case we resorted to a more complicated analysis via martingale inequalities.}
In particular, we sometimes find it useful to work in the probability space obtained by conditioning on the outcome of the exploration phase. Specifically, the \emph{\condSpace} is the probability space obtained by conditioning on the following events: that the exploration phase ends, that it has a specific duration $\TimeX$, and that arm $1$ is chosen for exploitation.

Throughout the analysis, we will obtain concentration bounds that hold with probability at least $1-2n^{-4}$. We will often take a Union Bound over all rounds $t$, which will imply success probability at least $1-2n^{-3}$. To simplify presentation, we will allow a slight abuse of notation: we will say \emph{with high probability} (abbreviated \emph{w.h.p.}), which will mean mean with probability at least $1-2n^{-3}$ or at least $1-2n^{-4}$, depending on the context.

To parameterize the algorithm, let us fix some $\chernoff = 12\,\ln(n)$ such that Theorem~\ref{thm:our-chernoff} with $C = \chernoff$ ensures success probability at least $1-2\,n^{-4}$.

\OMIT{ 
We will apply Chernoff Bounds to various sums such as, for example,
    $\left( \sum_t g_{i,t}\, I_{i,t} \right)$,
where each summand corresponds to a particular round $t$ and the sum is taken over some subset of rounds. In all our applications the summands are mutually independent because of ``fresh'' randomness from the random rewards (in the stochastic model) and/or internal randomization in the algorithm.

However, many of the sums that we encounter in our analysis do \emph{not} have independent summands. One reason is that a $t$-th round summand may depend on the arm selected by the algorithm in this round, which in turn may depend on the past history. Moreover, we will need to prove concentration inequalities conditional on some events in the execution of the algorithm. For these two reasons, setting up the sums so that Chernoff Bounds can be applied is a major issue in the analysis; our algorithm is designed with this issue in mind.%
} 

\OMIT{
Chernoff Bounds apply to sums of random variables that are independent. In general, independence can be violated if one conditions on various events in the algorithm, such as the events that the exploration phase ends at a given time $\TimeX$ and that arm $1$ is chosen for exploitation (see Section \ref{subsec:simpleSAO}). To ensure that Chernoff Bounds apply we will need to carefully define the random variables and/or choose appropriate conditioning. This issue is a relatively minor technical obstacle in this section. It becomes more serious for the $K$-armed algorithm in Section~\ref{sec:K-arms}, where using Chernoff Bounds seems no longer feasible and we are forced to apply more sophisticated martingale inequalities.
}

\subsection{Analysis: adversarial model}
\label{subsec:2arms-adversarial}

We need to analyze our algorithm in two different reward models. We start with the adversarial model, so that we can re-use some of the claims proved here to analyze the stochastic model.

Recall that $\TimeX$ denotes the duration of the exploration phase (which in general is a random variable). Following the convention from Section~\ref{subsec:simpleSAO} that whenever the exploration phase ends, the arm chosen for exploitation is arm $1$. (Note that we do not assume that arm $1$ is the best arm.)

We start the analysis by showing that the re-sampling schedule in the exploitation phase does not result in playing arm $2$ too often.

\begin{claim}\label{cl:P2-arm2}
During the exploitation phase, arm $2$ is played at most $O(\TimeX \log n)$ times w.h.p..
\end{claim}
\begin{proof}
We will work in the \condSpace{} $\cS$. We need to bound from above the sum $\sum_t I_{2,t}$, where $t$ ranges over the exploitation phase. However, Chernoff Bounds do not immediately apply since the number of summands itself is a random variable. Further, if we condition on a specific duration of exploitation then we break independence between summands. We sidestep this issue by considering an alternative algorithm in which exploitation lasts indefinitely (i.e., without the stopping conditions), and which uses the same randomness as the original algorithm. It suffices to bound from above the number of times that arm 2 is played during the exploitation phase in this alternative algorithm; denote this number by $N$. Letting $J_t$ be the arm selected in round $t$ of the alternative algorithm, we have that
    $N = \sum_{t=\TimeX+1}^n\; \ds1_{\{J_t=i\}}$
is a sum of 0-1 random variables, and in $\cS$ these variables are independent. Moreover, in $\cS$ it holds that
\begin{align*}
\E[N] = \textstyle{\sum_{t=\TimeX+1}^n}\, p_{2,t}
    = \TimeX\, \textstyle{\sum_{t=\TimeX+1}^n}\,\tfrac{1}{2t}
    = O(\TimeX \log n).
\end{align*}
Therefore, the claim follows from Chernoff Bounds.
\end{proof}

Now we connect the estimated cumulative rewards $\estimate{G}_{i,t}$ with the benchmark $G_{i,t}$. More specifically, we will bound from above several expressions of the form
    $|\estimate{H}_{i,t} - H_{i,t}|$.
Naturally, the upper bound for arm $1$ will be stronger since this arm is played more often during exploitation.  To ensure that the bound for arm $2$ is strong enough we need to play this arm ``sufficiently often" during exploitation. (Whereas Claim~\ref{cl:P2-arm2} ensures that we do not play it ``too often".) Here and elsewhere in this analysis, we find it more elegant to express some of the claims in terms of the average cumulative rewards (such as $H_{i,t}$, etc.)

\begin{claim}\label{cl:unbiased-estimate}~
\begin{OneLiners}
\item[(a)] With high probability,
    $|\estimate{H}_{i,\TimeX} - H_{i,\TimeX}| < 2\,\chernoff/\sqrt{\TimeX}$
    for each arm $i$.

\item[(b)] For any round $t$ in the exploitation phase, with high probability it holds that
\begin{align}\label{eq:cl:unbiased-estimate}
\begin{cases}
|\estimate{H}_{1,t}-H_{1,t}| < 3\, \chernoff /\sqrt{t},\\
|\estimate{H}_{2,t}-H_{2,t}| < 3\,\chernoff /\sqrt{\TimeX}.
\end{cases}
\end{align}
\end{OneLiners}
\end{claim}

\begin{proof}
For part (a), we are interested in the sum
    $\sum_{t\leq \TimeX}\; g_{i,t}\; I_{i,t}$.
As in the proof of Claim~\ref{cl:P2-arm2}, Chernoff Bounds do not immediately apply since the number of summands $\TimeX$ is a random variable (and conditioning on a particular value of $\TimeX$ tampers with independence between summands). So let us consider an alternative algorithm in which the exploration phase proceeds indefinitely, without the stopping condition, and uses the same randomness as the original algorithm.%
\footnote{Note that this is not the same ``alternative algorithm'' as the one in the proof of Claim~\ref{cl:P2-arm2}.}
Let $J_t$ be the arm selected in round $t$ of this alternative algorithm, and define
    $A_{i,t} = \sum_{s=1}^t\, g_{i,t}\, \ds1_{\{J_t=i\}}$.
Then (when run on the same problem instance) both algorithms coincide for any $t\leq \TimeX$, so in particular $\estimate{G}_{i,t} = 2\,A_{i,t}$. Now, $A_{i,t}$ is the sum of bounded independent random variables with expectation $G_{i,t}/2$. Therefore by Chernoff Bounds w.h.p. it holds that
    $|A_{i,t} - G_{i,t}/2| < \chernoff \sqrt{t}$
for each $t$, which implies the claim.

For part (b), we will analyze the exploitation phase separately. Let us work in the \condSpace{} $\cS$. We will consider the alternative algorithm from the proof of Claim~\ref{cl:P2-arm2} (in which exploitation continues indefinitely). This way we do not need worry that we implicitly condition on the event that a particular round $t>\TimeX$ belongs to the exploitation phase. Clearly, it suffices to prove~\eqref{eq:cl:unbiased-estimate} for this alternative algorithm. To facilitate the notation, define the time interval
    $\INT = \{\TimeX+1, \ldots, t\}$,
and denote
    $G_{i,\INT} = \sum_{s\in \INT} g_{i,t}$
and
    $\estimate{G}_{i,\INT} = \sum_{s\in \INT} \estimate{g}_{i,t}$.

To handle arm $1$, note that in $\cS$, $\estimate{G}_{i,\INT}$ is a sum of independent random variables, with expectation $G_{i,\INT}$. Since $p_{1,t}\geq \tfrac12$ for any $t\in\INT$, the summands $\estimate{g}_{1,t}$ are bounded by 2. Therefore by Chernoff Bounds with high probability it holds that
\begin{align*}
|\estimate{G}_{1,\INT} - G_{1,\INT}| < 2\,\chernoff \sqrt{t-\TimeX}.
\end{align*}
From this and part (a) it follows that w.h.p.
\begin{align*}
|\estimate{G}_{1,t} - G_{1,t}|
    < 2\,\chernoff(\sqrt{\TimeX} + \sqrt{t-\TimeX})
    < 3\,\chernoff \sqrt{t},
\end{align*}
which implies the claim for arm $1$.

Handling arm $2$ requires a little more work since the summands $\estimate{g}_{2,t}$ may be large (since they have a small probability $p_{2,t}$ in the denominator). For each $t\in \INT$,
\begin{align*}
\estimate{G}_{2,\INT}
    &= {\textstyle \sum_{s\in \INT}}\;
        \frac{2s}{\TimeX} \; g_{2,s} \; I_{2,s}
    = \frac{2t}{\TimeX} \; {\textstyle\sum_{s\in \INT}}\;
        \frac{s}{t} \; g_{2,s} \; I_{2,s}
    = \frac{2t}{\TimeX} \; {\textstyle\sum_{s\in \INT}}\; X_s,
\end{align*}
where
    $X_s = \tfrac{s}{t}\; g_{2,s} \, I_{2,s} \in [0,1]$.
In $\cS$, random variables $X_s,\, s\in\INT$ are mutually independent, and the expectation of their sum is
\begin{align*}
\mu \triangleq \textstyle \bigE{ \sum_{s\in \INT}\, X_s }
    = \frac{\TimeX}{2t}\; \bigE{ \estimate{G}_{2,\INT}}
    = \frac{\TimeX}{2t}\; G_{2,\INT}
    \leq \frac{\TimeX}{2}\, \frac{t-\TimeX}{t}.
\end{align*}
Noting that $G_{2,\INT} \leq t-\TimeX$ and letting
    $\alpha = \tfrac{\TimeX}{t}$,
we obtain $\mu \leq \frac{\TimeX}{2} (1-\alpha)$.
By Chernoff Bounds w.h.p. it holds that
\begin{align*}
| \textstyle{\sum_{s\in \INT}}\, X_s - \mu| &< \chernoff
    \sqrt{\TimeX\,(1-\alpha)}.
\end{align*}
Going back to $\estimate{G}_{2,\INT}$ and $G_{2,\INT}$, we obtain:
\begin{align*}
| \estimate{G}_{2,\INT} - G_{2,\INT}|
    &< \frac{2t}{\TimeX}\; \chernoff \sqrt{\TimeX\,(1-\alpha)}
    < \chernoff\;  \frac{2t}{\sqrt{\TimeX}}\; \sqrt{1-\alpha}.
\end{align*}
From part (a), we have that
$|\estimate{G}_{i,\TimeX} - G_{i,\TimeX}|
    < \chernoff\, \tfrac{2t}{\sqrt{\TimeX}}\, \alpha$.
Therefore,
\begin{align*}
|\estimate{G}_{2,t} - G_{2,t}|
    < \chernoff \; \frac{2t}{\sqrt{\TimeX}}\;(\sqrt{\alpha} + \sqrt{1-\alpha})
    < \chernoff\;\frac{3t}{\sqrt{\TimeX}}. \qquad \qedhere
\end{align*}
\end{proof}

Combining Claim~\ref{cl:unbiased-estimate}(b) and Condition~\eqref{eq:cond1}, we obtain:

\begin{corollary}\label{cor:benchmark}
In the exploitation phase, for any round $t$ (except possibly the very last round in the phase) it holds w.h.p. that
    $G_{1,t}> G_{2,t}$.
\end{corollary}

\OMIT{ 
For $>2$ arms, Claim~\ref{cl:unbiased-estimate}(b) and Claim~\ref{cl:unbiased-estimate}(b) should be replaced by a single claim:
\begin{align}
|\estimate{H}_{j,t}-H_{j,t}| < 3\, \chernoff \log(t)/\sqrt{\tau_j}
\end{align}
for each arm $j$, where $\tau_j$ is time when arm $j$ is ``disqualified" by the algorithm.
} 

By Corollary~\ref{cor:benchmark}, regret accumulated by round $t$ in the exploitation phase is, with high probability, equal to
    $G_{1,t} - \algo{G}_{1,t} - \algo{G}_{2,t}$.
The following claim upper-bounds this quantity by $O(\sqrt{t}\,\log^2 n)$.
The proof of this claim contains our main regret computation.

\begin{claim}\label{cl:final-computation}
For any round $t$ in the exploitation phase it holds w.h.p. that
\begin{align*}
\algo{G}_{1,t} + \algo{G}_{2,t} - G_{1,t}
    \geq - O(\sqrt{t}\,\log^2 n).
\end{align*}
\end{claim}

\begin{proof}
Throughout this proof, let us assume that the high-probability events in Claim~\ref{cl:P2-arm2} and Claim~\ref{cl:unbiased-estimate}(b) actually hold; we will omit ``with high probability'' from here on.

Let $t$ be some (but not the last) round in the exploitation phase.
First,
\begin{align}\label{eq:payoff-arm1}
\algo{H}_{1,t} - H_{1,t}
    = \left[ \estimate{H}_{1,t}-H_{1,t} \right] +
            \left[ \algo{H}_{1,t} - \estimate{H}_{1,t} \right]
    \geq  - O(\chernoff / \sqrt{t}).
\end{align}
We have upper-bounded the two square brackets in~\eqref{eq:payoff-arm1} using, respectively, Claim~\ref{cl:unbiased-estimate}(b) and Condition~\eqref{eq:cond2}. We proved that  algorithm's average for arm $1$ ($\algo{H}_{1,t}$) is not too small compared to the corresponding benchmark average $H_{1,t}$, and we used the estimate $\estimate{H}_{1,t}$ as an intermediary in the proof.

Similarly, using
Condition~\eqref{eq:cond1},
Condition~\eqref{eq:cond2},
and Claim~\ref{cl:unbiased-estimate}(b) to upper-bound the three square brackets in the next equation, we obtain that
\begin{align}
\algo{H}_{2,t} - H_{1,t}
    = \left[ \estimate{H}_{2,t}- \estimate{H}_{1,t} \right] +
          \left[ \algo{H}_{2,t} - \estimate{H}_{2,t} \right] +
          \left[ \estimate{H}_{1,t}- H_{1,t} \right]
    \geq  - O(\chernoff/ \sqrt{\TimeX}). \label{eq:payoff-arm2}
\end{align}
Here we have proved that the algorithm did not do too badly playing arm $2$, even though this arm was supposed to be suboptimal. Specifically, we establish that algorithm's average
for arm $2$ ($\algo{H}_{2,t}$) is not too small compared to the benchmark average \emph{for arm 1} ($H_{1,t}$). Again, the estimates $\estimate{H}_{1,t}$ and $\estimate{H}_{2,t}$ served us as intermediaries in the proof.

Finally, let us go from bounds on average rewards to bounds on cumulative rewards (and prove the claim).
Combining~\eqref{eq:payoff-arm1},~\eqref{eq:payoff-arm2} and Claim~\ref{cl:P2-arm2}, we have:
\begin{align}
\algo{G}_{1,t} + \algo{G}_{2,t} - G_{1,t}
    &= \textstyle{\sum_{i=1,2}} \;
        T_j(t) \left[ \algo{H}_{i,t} - H_{1,t} \right] \nonumber \\
    &\geq - O(\chernoff) \left[\,
            T_1(t) /\sqrt{t} + T_2(t) /\sqrt{\TimeX}
        \,\right] \nonumber \\
    &\geq - O(\chernoff) (\sqrt{t} + \sqrt{\TimeX} \log n) \nonumber \\
    &\geq - O(\sqrt{t}\,\log^2 n) . \qedhere
\end{align}
\end{proof}

Now we are ready for the final computations. We will need to consider three cases, depending on which phase the algorithm is in when it halts (i.e., reaches the time horizon).

First, if the exploration phase never ends then by Claim~\ref{cl:unbiased-estimate}(a) w.h.p. it holds that
    $|\estimate{H}_{i,n}-H_{i,n}| < 2\, \chernoff /\sqrt{n}$
for each arm $i$, and the exit condition~\eqref{eq:cond-exploration} never fails. This implies the claimed regret bound
    $R_n \leq O(\sqrt{n} \log n)$.

From here on let us assume that the exploration phase ends at some $\TimeX<n$. Define regret on the time interval $[a,b]$ as
\begin{align*}
R_{[a,b]} &= \max_{i\in \{1,2\}}
    \textstyle{\sum_{a=1}^b\; g_{i,t}}
    - \textstyle{\sum_{s=a}^b\; g_{I_t,t}}.
\end{align*}
Let $t$ be the last round in the exploitation phase.
By Corollary~\ref{cor:benchmark} and Claim~\ref{cl:final-computation} we have
\begin{align*}
R_{[1,t-1]}
    &=  G_{1,t}- \algo{G}_{1,t} - \algo{G}_{2,t}  \leq O(\sqrt{n}\,\log^2 n).
\end{align*}
If $t=n$ (i.e., the algorithm halts during exploitation) then we are done.

Third, if the algorithm enters the adversarial phase then we can use the regret bound for $\EXPp$ in~\cite{bandits-exp3}, which states that w.h.p.
    $R_{[t,n]} \leq O(\sqrt{n})$.
Therefore
\begin{align*}
    R_n \leq R_{[1,t-1]} + R_{[t,n]} \leq O(\sqrt{n}\,\log^2 n).
\end{align*}
This completes the proof of Theorem~\ref{thm:2-arms}(a).

\subsection{Analysis: stochastic model}
\label{subsec:2arms-stochastic}

We start with a simple claim that w.h.p. each arm is played sufficiently often during exploration, and arm $1$ is played sufficiently often during exploitation. This claim complements Claim~\ref{cl:P2-arm2} (which we will also re-use) which states that arm $2$ is not played too often during exploitation.

\begin{claim}\label{cl:P2-arms}
With high probability it holds that:
\begin{OneLiners}
\item[(a)] during the exploration phase, each arm is played at least $\TimeX/4$ times.
\item[(b)] during the exploitation phase, $T_1(t)\geq t/4$ for each time $t$.
\end{OneLiners}
\end{claim}
\begin{proof}
Both parts follow from Chernoff Bounds. The only subtlety is to ensure that we do not condition the summands (in the sum that we apply the Chernoff Bounds to) on a particular value of $\TimeX$ or on the fact that arm $1$ is chosen for exploitation.

For part (a), without loss of generality assume that $n$ fair coins are tossed in advance, so that in the $t$-th round of exploration we use the $t$-th coin toss to decide which arm is chosen. Then by Chernoff Bounds for each $t$ w.h.p. it holds that among the first $t$ coin tosses there are at least
    $t/2 - \chernoff \sqrt{t/2}$
heads and at least this many tails. We take the Union Bound over all $t$, so in particular this holds for $t=\TimeX$. Therefore w.h.p. we have:
\begin{align}\label{eq:cl:P2-arms}
    T_i(\TimeX)\geq \TimeX/2 - \chernoff \sqrt{\TimeX/2}.
\end{align}
The claim follows from~\eqref{eq:cl:P2-arms} because we force exploration to last for at least $\Omega(\chernoff^2)$ rounds.

For part (b), let us analyze the exploitation phase separately. We are interested in the sum $\sum_s I_{1,s}$, where $s$ ranges over all rounds in the exploitation phase. We will work in the \condSpace. The indicator variables $I_{1,s}$, for all rounds $s$ during exploitation, are mutually independent. Therefore Chernoff Bounds apply, and w.h.p.
\begin{align*}
    T_1(t) - T_1(\TimeX) \geq (t-\TimeX)/2 - \chernoff \sqrt{t-\TimeX}.
\end{align*}
Using~\eqref{eq:cl:P2-arms}, it follows that
    $T_1(t) \geq t/2 - \chernoff (\sqrt{\TimeX}+\sqrt{t-\TimeX})
            \geq t/2 - \chernoff \sqrt{t} \geq t/4$.
\end{proof}


Recall that Claim~\ref{cl:unbiased-estimate}(b) connects algorithm's estimate $\estimate{H}_{i,t}$ and the benchmark average $H_{i,t}$ (we will re-use this claim later in the proofs). In the stochastic model these two quantities, as well as the algorithm's average $\algo{H}_{i,t}$, are close to the respective expected reward $\mu_i$.  The following lemma makes this connection precise.

\begin{claim}\label{cl:arms-stoch}
Assume the stochastic model. Then during the exploitation phase for each arm $i$ and each time $t$  the following holds with high probability:
\begin{align*}\begin{cases}
    |H_{i,t} - \mu_i| \leq \chernoff /\sqrt{t},\\
    |\algo{H}_{1,t} - \mu_1| \leq 2\,\chernoff /\sqrt{t}, \\
    |\algo{H}_{2,t} - \mu_2| \leq 2\,\chernoff /\sqrt{\TimeX}.
\end{cases}\end{align*}
\end{claim}

\begin{proof}
All three inequalities follow from Chernoff Bounds. The first inequality follows immediately. To obtain the other two inequalities, we claim that w.h.p. it holds that
\begin{align}\label{eq:arms-stoc}
|\algo{H}_{i,t} - \mu_i| \leq \chernoff /\sqrt{T_i(t)}.
\end{align}
Indeed, note that without loss of generality $T$ independent samples from the reward distribution of arm $i$ are drawn in advance, and then the reward from the $\ell$-th play of arm $i$ is the $\ell$-th sample. Then by Chernoff Bounds the bound~\eqref{eq:arms-stoc} holds w.h.p. for each $T_i(t)=l$, and then one can take the Union Bound over all $l$ to obtain~\eqref{eq:arms-stoc}. Claim proved.

Finally, we use~\eqref{eq:arms-stoc} and plug in the lower bounds on $T_i(t)$ from Claim~\ref{cl:P2-arms}(ab).
\end{proof}

Now that we have all the groundwork, let us argue that in the stochastic model the consistency condition in the algorithm are satisfied with high probability.

\begin{corollary}\label{cor:arms-stoch}
Assume the stochastic model. Then in each round $t$ of the exploitation phase, with high probability the following holds:
\begin{align}\label{eq:cor-arms-stoch}
16 \,\chernoff/\sqrt{\TimeX}
    \leq \mu_1 - \mu_2
    \leq 32\, \chernoff/\sqrt{\TimeX}.
\end{align}
Moreover, conditions
    (\ref{eq:cond1}-\ref{eq:cond2})
are satisfied.
\end{corollary}
\begin{proof}
Condition~\eqref{eq:cond2} follows simply by combining Claim~\ref{cl:unbiased-estimate}(b) and Claim~\ref{cl:arms-stoch}.

To obtain~\eqref{eq:cor-arms-stoch}, we note that by Claim~\ref{cl:unbiased-estimate}(b) and Claim~\ref{cl:arms-stoch} w.h.p. it holds that
\begin{align}\label{eq:cor-arms-stoch-2}
|\estimate{H}_{1,t} - \mu_1| + |\estimate{H}_{2,t} - \mu_2|
    \leq 8\,\chernoff /\sqrt{\TimeX}.
\end{align}
Recall that Condition~\eqref{eq:cond-exploration} holds at time $t= \TimeX-1$, and fails at $t= \TimeX$. This in conjunction with~\eqref{eq:cor-arms-stoch-2} (for $t=\TimeX$) implies~\eqref{eq:cor-arms-stoch}. In turn,~\eqref{eq:cor-arms-stoch} with~\eqref{eq:cor-arms-stoch-2} imply Condition~\eqref{eq:cond1}.
\end{proof}

To complete the proof of Theorem~\ref{thm:2-arms}(b), assume we are in the stochastic model with gap $\Delta = |\mu_1-\mu_2|$. In the rest of the argument, we omit ``with high probability". If the exploration phase never ends, it is easy to see that
    $\Delta\leq O(1/\sqrt{n})$,
and we are done since trivially
    $\oR_n \leq \Delta\, n \leq O(\tfrac{1}{\Delta})$.
Else,  by Corollary~\ref{cor:arms-stoch} it holds that arm $1$ is optimal,
    $\TimeX = \Theta(\chernoff/ \Delta)^2$
and moreover that the exploitation phase never ends. Now, by Claim~\ref{cl:P2-arm2} in the exploitation phase the suboptimal arm $2$ is played at most $O(\TimeX \log n)$ times. Therefore $\oR_n \leq O(\tfrac{1}{\Delta}\,\log^3 n)$.

\section{The {\alg} algorithm for the general case} \label{sec:K-arms}

In this section we treat the general case: $K$ arms and adaptive adversary. The proposed algorithm {\alg} (Stochastic and Adversarial Optimal), is described precisely in Algorithm~\ref{alg:soa} (see page~\pageref{alg:soa}). On a high-level, {\alg} proceeds similarly to the simplified version in Section \ref{sec:2-arms}, but there are a few key differences.

First, the exploration and exploitation phases are now interleaved. Indeed, {\alg} starts with all arms being ``active'', and then it successively ``deactivates'' them as they turn out to be suboptimal. Thus, the algorithm evolves from pure exploration (when all arms activated) to pure exploitation (when all arms but the optimal one are deactivated).

Second, in order to make the above evolution smooth we adopt a more complicated (re)sampling schedule that the one we used in Section \ref{sec:2-arms}. Namely, the probability of selecting a given arm continuously increases while this arm stays active, and then continuously decreases when it gets deactivated, and the transition between the two phases is also continuous. For the precise equation, see Equation~\eqref{eq:Karm-algo-update} in Algorithm 1.

Third, this more subtle behavior of the (re)sampling probabilities $p_{i,t}$ in turn necessitates more complicated consistency conditions (e.g. see Condition~\eqref{eq:test1} compared to Condition~\eqref{eq:cond2}), and a more intricate analysis. The key in the analysis is to obtain the good concentration properties of the different estimators, which we accomplish by exhibiting martingale sequences and resorting to Bernstein's inequality for martingales (Theorem \ref{th:be}).

\newcommand{\minTau}{t^*_i} 

\begin{algorithm}[t]
\begin{algorithmic}[1]
\State $A \gets \{1,\hdots,K\}$ \Comment{$A$ is the set of active arms}
\For{$i=1,\hdots,K$} \Comment{Initialization}
\State $\tau_i \gets n$ \Comment{$\tau_i$ is the time when arm $i$ is deactivated}
\State $p_i \gets 1/K$ \Comment{$p_i$ is the probability of selecting arm $i$}
\EndFor
\For{$t=1,\hdots,n$} \Comment{Main loop}
\State Play $I_t$ at random from $p$ \Comment{Selection of the arm to play}
\For{$i=1,\hdots,K$} \Comment{Test of four properties for arm $i$}
\If{  \Comment{Test if arm $i$ should be deactivated}
\begin{equation} \label{eq:test2}
 i \in A, \; \text{and} \; \max_{j \in A} \tilde{H}_{j,t} - \tH >
6 \sqrt{\frac{4 K \log(\beta)}{t} +
5\left( \frac{K \log(\beta)}{t} \right)^2}
\end{equation}
\State
}
{$A \gets A \setminus \{i\}$, $\tau_i \gets t$ and $q_i \gets p_i$} \Comment{Deactivation of arm $i$}
\EndIf
\Comment{$q_i$ denotes the probability of arm $i$
at the moment when it was de-activated}

\If{ {\bf one of the three following properties is satisfied} \State}
{ Start Exp3.P with the parameters described in [Theorem 2.4, \cite{Bub10}]}
\\  \Comment{Test if stochastic model still valid for arm $i$}
\\ \Comment{First, test if the two estimates of $H_{i,t}$ are consistent;
    let $\minTau = \min(\tau_i, t)$.}
\begin{align} \label{eq:test1}
\left|\tH - \hH \right|
    > \sqrt{\frac{2 \log(\beta)}{T_i(t)}}
        + \sqrt{4 \left(\frac{K \minTau}{t^2} + \frac{t-\minTau}{q_i \tau_i t} \right) \log(\beta) +
5 \left( \frac{K \log(\beta)}{\minTau} \right)^2}.
\end{align}
\\ \Comment{Second, test if the estimated suboptimality of arm $i$ did not increase too much}
\begin{equation}
\label{eq:test3}
i \not\in A, \; \text{and} \; \max_{j \in A} \tilde{H}_{j,t} - \tH >
10 \sqrt{\frac{4 K \log(\beta)}{\tau_i-1} +
5 \left( \frac{K \log(\beta)}{\tau_i-1} \right)^2}.
\end{equation}
\\ \Comment{Third, test if arm $i$ still seems significantly suboptimal}
\begin{equation}
\label{eq:test4}
i \not\in A, \; \text{and} \;
\max_{j \in A} \tilde{H}_{j,t} - \tH \leq
2 \sqrt{\frac{4 K \log(\beta)}{\tau_i} +
5 \left( \frac{K \log(\beta)}{\tau_i} \right)^2}.
\end{equation}

\EndIf
\EndFor  \Comment{End of testing}

\For{$i=1,\hdots,K$} \Comment{Update of the probability of selecting arm $i$}
\begin{align}\label{eq:Karm-algo-update}
p_i \gets \frac{q_i \; \tau_i}{t+1}\; \ds1_{i \not\in A}
    + \frac{1}{|A|} \left(1 - \sum_{j \not\in A} \frac{q_j \; \tau_j}{t+1} \right) \ds1_{i \in A}.
\end{align}
\EndFor
\EndFor
\end{algorithmic}
\caption{The {\alg} strategy with parameter $\beta > 1$} \label{fig:alg}
\label{alg:soa}
\end{algorithm}

Recall that the crucial parameter for the stochastic model is the \emph{minimal gap}
    $    \Delta = \min_{i:\, \Delta_i>0} \Delta_i$,
where
    $\Delta_i = (\max_{1 \leq j \leq K} \mu_j) - \mu_i$
is the \emph{gap} of arm $i$.
Our main result is formulated as follows:

\begin{theorem} \label{th:1}
{\alg} with $\beta=n^4$ satisfies
\[\left\{ \begin{array}{rcll}
\E[\oR_n]  &\leq& O \left( \frac{K \log(K) \log^2(n)}{\Delta} \right)
    & \text{in the stochastic model}, \\
\E[ R_n ] &\leq& O \left( \log(K) \log^{3/2}(n) \sqrt{n K} \right)
    &\text{in the adversarial model}.
\end{array}\right. \]
More precisely, for any $\delta \in (0,1)$, with probability at least $1-\delta$, {\alg} with $\beta=10 K n^3 \delta^{-1}$ satisfies in the stochastic model:
$$\oR_n \leq\frac{260 K (1+ \log K) \log^2(\beta)}{\Delta},$$
and in the adversarial model:
$$R_n \leq 60 (1+\log K) (1+\log n) \sqrt{n K \log(\beta) + 5 K^2 \log^2(\beta)} + 200 K^2 \log^2(\beta).$$
\end{theorem}

\clearpage

We divide the proof into three parts. In Section \ref{sec:proof1} we propose several concentration inequalities for the different quantities involved in the algorithm.
Then we make a deterministic argument conditional on the event that all these concentration inequalities hold true. First, in Section \ref{sec:proof2}, we analyze stochastic rewards, and Section \ref{sec:proof3} concerns the adversarial rewards.

Let us discuss some notation. Recall that we denote by $p_{i,t}$ the probability that the algorithm selects arm $i$ at time $t$; this probability is denoted by $p_i$ in the description of the algorithm. As in Algorithm~\ref{alg:soa}, $q_i$ will denote the probability of arm $i$ at the moment when this arm was deactivated. Let $A_t$ denote the set of active arms at the end of time step $t$.
We also introduce $\tau_0$ as the last time step before we start $\EXPp$, with a convention that
$\tau_0=n$ if we never start $\EXPp$. Moreover note that with this notation, if
$\tau_i < \tau_0$ then we have $q_i = p_{i, \tau_i}$.
We generalize this notation and set $q_i : = p_{i,\; \min(\tau_i,\tau_0)}$. For sake of notation, in the following $\tau_i$ denotes the minimum between the time when arm $i$
is deactivated and the last time before we start $\EXPp$, that is
    $\tau_i \gets \min(\tau_i, \tau_0)$.

\subsection{Concentration inequalities}
\label{sec:proof1}

We start with two standard concentration inequalities for martingale sequences.

\begin{theorem}[Hoeffding-Azuma's inequality for martingales, \cite{Hoe63}] \label{th:ha}~\\
Let $\cF_1 \subset \dots \subset \cF_n$ be a filtration, and $X_1,\dots,X_n$ real random variables such that
$X_t$ is $\cF_t$-measurable, $\E(X_t|\cF_{t-1})=0$ and $X_t \in [A_t, A_t + c_t]$ where $A_t$ is a random variable $\cF_{t-1}$-measurable and $c_t$ is a positive constant. Then, for any $\epsilon>0$, we have
  \begin{equation} \label{eq:ha1}
  \P\Big( \sum_{t=1}^n X_t \ge \epsilon \Big) \le \exp\left( - \frac{2 \epsilon^2}{\sum_{t=1}^n c_t^2} \right) ,
  \end{equation}
or equivalently for any $\delta>0$,
with probability at least $1-\delta$, we have
  \begin{equation} \label{eq:ha2}
  \sum_{t=1}^n X_t \le \sqrt{\frac{\log(\delta^{-1})}{2} \sum_{t=1}^n c_t^2}.
  \end{equation}
\end{theorem}

\begin{theorem}[Bernstein's inequality for martingales, \cite{Fre75}] \label{th:be}~\\
Let $\cF_1 \subset \dots \subset \cF_n$ be a filtration, and $X_1,\dots,X_n$ real random variables such that
$X_t$ is $\cF_t$-measurable, $\E(X_t|\cF_{t-1})=0$, $|X_t|\le b$ for some $b>0$ and let $V_n = \sum_{t=1}^n \E(X_t^2|\cF_{t-1})$.
Then, for any $\epsilon>0$, we have
  \begin{equation} \label{eq:be1}
  \P\Big( \sum_{t=1}^n X_t \ge \epsilon \; \text{and} \; V_n \leq V\Big) \le \exp\left( - \frac{\epsilon^2}{2 V + 2 b \epsilon / 3} \right) ,
  \end{equation}
and for any $\delta>0$,
with probability at least $1-\delta$, we have either $V_n > V$ or
  \begin{equation} \label{eq:be2}
  \sum_{t=1}^n X_t \le \sqrt{2 V \log(\delta^{-1})} + \frac{b\log(\delta^{-1})}{3} .
  \end{equation}
\end{theorem} 

Next we derive a version of Bernstein's inequality that suits our needs.

\begin{lemma} \label{lem:be3}
Let $\cF_1 \subset \dots \subset \cF_n$ be a filtration, and $X_1,\dots,X_n$ real random variables such that
$X_t$ is $\cF_t$-measurable, $\E(X_t|\cF_{t-1})=0$ and $|X_t|\le b$ for some $b>0$. Let $V_n = \sum_{t=1}^n \E(X_t^2|\cF_{t-1})$ and $\delta > 0$. Then
with probability at least $1 - \delta$,
$$\sum_{t=1}^n X_t \le \sqrt{4 V_n \log(n \delta^{-1}) + 5 b^2 \log^2(n \delta^{-1})}.$$
\end{lemma}

\begin{proof}
The proof follows from Theorem \ref{th:be} along with an union bound on the events
$V_n \in [x, x + b]$, $x \in \{0, b^2, 2 b^2, \hdots, (n-1) b^2\}$. It also uses $\sqrt{a} + \sqrt{b} \leq \sqrt{2 (a + b)}$.
\end{proof}

Now let us use this martingale inequality to derive the concentration bound for (average) estimated cumulative rewards $\tH$. Recall that $\tH$ is an estimator of $\oH$, so we want to upper-bound the difference $|\tH-\oH|$, and in the stochastic model $\tH$ is an estimator of the true expected reward $\mu_i$, so we want to upper-bound the difference $|\tH-\mu_i|$.

\begin{lemma} \label{lem:1}
For any arm $i \in \{1,\hdots,K\}$ and any time $t \in \{1,\hdots,n\}$, in the stochastic model we have with probability at least $1-\delta$, if $t \leq \tau_0$,
$$\left| \tH - \mu_i \right| \leq
\sqrt{4 \left(\frac{K \min(\tau_i,t)}{t^2} + \frac{\max(t - \tau_i, 0)}{q_i \tau_i t}\right) \log(2 t^2 \delta^{-1}) +
5 \left( \frac{K \log(2 t^2 \delta^{-1})}{\min(\tau_i, t)} \right)^2} .$$
Moreover in the adversarial model we have with probability at least $1-\delta$, if $t \leq \tau_0$,
$$\left| \tH - \oH \right| \leq
\sqrt{4 \left(\frac{K \min(\tau_i,t)}{t^2} + \frac{\max(t - \tau_i, 0)}{q_i \tau_i t}\right) \log(2 t^2 \delta^{-1}) +
5 \left( \frac{K \log(2 t^2 \delta^{-1})}{\min(\tau_i, t)} \right)^2
} .$$
\end{lemma}

\begin{proof}
The proof of the two concentration inequalities is similar, so we restrict our attention to the adversarial model.
Let $(\cF_s)$ be the filtration associated to the historic of the strategy. We introduce the following sequence of independent random variables:
for $1 \leq i \leq K$, $1 \leq s \leq n$ and $p \in [0,1]$, let $Z_{s}^i(p) \sim \Ber(p)$.  Then for $t \leq \tau_0$ we have,
$$\tG = \sum_{s=1}^t g_{i,s} \left(\frac{Z_s^i(p_{i,s})}{p_{i,s}} \ds1_{s \leq \tau_i} + \frac{s}{q_i \tau_i} Z_s^i\left(\frac{q_i \tau_i}{s}\right)
\ds1_{s > \tau_i} \right) .$$

For $T \in \{1,\hdots,n\}$, let
$$ X_s^i(T) = \left(\frac{Z_s^i(p_{i,s})}{p_{i,s}} -1\right) g_{i,s} \ds1_{s \leq \tau_i \leq T} +
\left(\frac{s}{q_i T} Z_s^i\left(\frac{q_i T}{s}\right) -1\right) g_{i,s} \ds1_{s > T \geq \tau_i}.$$ We have, for $t\leq \tau_0$,
$$\tG - \G = \sum_{s=1}^t X_s^i(\tau_i).$$
Now remark that $(X_s^i(T))_{1 \leq s \leq t}$ is a martingale difference sequences such that $|X_s^i(T)| \leq K \max\left(\frac{t}{T},1\right)$
(since $p_{i,s} \geq 1/K$ when $s \leq \tau_i$) and
$$\sum_{s=1}^t \E \left( (X_s^i(T))^2 | \cF_{s-1} \right) \leq \sum_{s=1}^{\min(\tau_i,t)} \frac{1}{p_{i,s}} + \frac{t \max(t - T, 0)}{q_i T}.$$
Thus, using Lemma \ref{lem:be3},
we obtain that with probability at least $1 - \delta$,
$$\sum_{s=1}^t X_{s}^i(T) \leq \sqrt{4 \left(\sum_{s=1}^{\min(\tau_i,t)} \frac{1}{p_{i,s}} + \frac{t \max(t - T, 0)}{q_i T}\right) \log(t \delta^{-1}) + 5 K^2 \max\left(\left(\frac{t}{T}\right)^2,1\right)
\log^2(t \delta^{-1})} .$$
Then, using an union bound over $T$, we obtain the claimed inequality by taking $T=\tau_i$ (with another union bound to get the two-sided inequality).
\end{proof}

Next, we analyze the (average) cumulative reward $\hH$ collected by the algorithm. Again, in the stochastic model $\hH$ can be used as an estimate of the true expected reward $\mu_i$, and it is not hard to see that it is a reasonably sharp estimate.

\begin{lemma} \label{lem:2}
For any arm $i \in \{1,\hdots,K\}$, in the stochastic model we have with probability at least $1-\delta$, for any time $t \in \{1,\hdots,n\}$,
if $T_i(t) \geq 1$,
$$\left| \hH - \mu_i \right| \leq
\sqrt{\frac{2 \log( 2 n \delta^{-1} )}{T_i(t)}}.$$
\end{lemma}

\begin{proof}
This follows via an union bound over the value of $T_i(t)$ and a standard Hoeffding's inequality for independent random variables, see Theorem \ref{th:ha}.
\end{proof}

Next we show that, essentially,
    $T_i(t) \leq \tilde{O}(q_i \tau_i + \sqrt{q_i \tau_i}) $.

\begin{lemma} \label{lem:3}
For any $i \in \{1,\hdots,K\}, t \in \{1,\hdots,n\}$, with probability at least $1-\delta$, if $t \leq \tau_0$,
$$T_i(t) \leq q_i \tau_i (1 + \log t)
+ \sqrt{4 q_i \tau_i (1 + \log t) \log(t \delta^{-1}) + 5 \log^2(t \delta^{-1})}.$$
\end{lemma}

\begin{proof}
Using the notation of the proof of Lemma \ref{lem:1}, we have for $t \leq \tau_0$,
$$T_i(t) = \sum_{s=1}^t Z_s^i(p_{i,s}) \ds1_{s \leq \tau_i} + Z_s^i\left(\frac{q_i \tau_i}{s}\right) \ds1_{s > \tau_i}.$$
Let $$X_s^i = (Z_s^i(p_{i,s}) - p_{i,s}) \ds1_{s \leq \tau_i} + \left(Z_s^i\left(\frac{q_i \tau_i}{s}\right) - \frac{q_i \tau_i}{s}\right)
\ds1_{s > \tau_i}.$$
Then $(X_s^i)$ is a martingale difference
sequence such that $|X_s^i| \leq 1$ and, since $p_{i,s}$ is increasing in $s$ for $s\leq \tau_i$, it follows that
$$\sum_{s=1}^t \E ( (X_s^i)^2 | \cF_{s-1} ) \leq q_i \tau_i + \sum_{s=\tau_i+1}^t \frac{q_i \tau_i}{s} \leq q_i \tau_i (1+\log t) .$$
Thus using Lemma \ref{lem:be3} we obtain that with probability at least $1-\delta$:
$$\sum_{s=1}^t X_s^i(T) \leq \sqrt{4 q_i \tau_i (1 + \log t ) \log(t \delta^{-1}) + 5 \log^2(t \delta^{-1})}.$$
It implies that
$$\sum_{s=1}^t Z_s^i(p_{i,s}) \ds1_{s \leq \tau_i} + Z_s^i\left(\frac{q_i \tau_i}{s}\right) \ds1_{s > \tau_i} \leq q_i \tau_i (1 + \log t )
+ \sqrt{4 q_i \tau_i (1 + \log t ) \log(t \delta^{-1}) + 5 \log^2(t \delta^{-1})},$$
which is the claimed inequality.
\end{proof}

The next lemma restates regret guarantee for $\EXPp$ in terms of our setting. Instead of using the original guarantee from \cite{bandits-exp3}, we take an improved bound from \cite{Bub10} (namely, Theorem 2.4 in \cite{Bub10}).

\begin{lemma} \label{lem:4}
In the adversarial model, with probability at least $1-\delta$, we have
$$\max_{i \in \{1,\hdots,K\}} \sum_{t=\tau_0+1}^n g_{i,t} - \sum_{t=\tau_0+1}^n g_{I_t,t} \leq 5.15 \sqrt{(n-\tau_0) K \log(K \delta^{-1})}.$$
\end{lemma}

Let $\beta= 10 K n^3 \delta^{-1}$. Putting together the results of Lemma \ref{lem:1}, \ref{lem:2}, \ref{lem:3} and \ref{lem:4},
we obtain that with probability at least $1-\delta$, the following inequalities hold true
for any arm $i \in \{1,\hdots,K\}$ and any time $t \in \{1,\hdots,\tau_0\}$:
\begin{align}
& \text{In the stochastic model,} \notag\\
& \left| \tH - \mu_i \right| \leq
\sqrt{4 \left(\frac{K \min(\tau_i,t)}{t^2} + \frac{\max(t - \tau_i, 0)}{q_i \tau_i t}\right) \log(\beta) +
5 \left( \frac{K \log(\beta)}{\min(\tau_i, t)} \right)^2}. \label{eq:conc1} \\
& \text{In the adversarial model,} \notag\\
& \left| \tH - \oH \right| \leq
\sqrt{4 \left(\frac{K \min(\tau_i,t)}{t^2} + \frac{\max(t - \tau_i, 0)}{q_i \tau_i t}\right) \log(\beta) +
5 \left( \frac{K \log(\beta)}{\min(\tau_i, t)} \right)^2}. \label{eq:conc2} \\
& \text{In the stochastic model,} \notag \\
     &\left| \hH - \mu_i \right| \leq
\sqrt{\frac{2 \log(\beta)}{T_i(t)}}. \label{eq:conc3} \\
& \text{In both models,} \notag\\
& T_i(t) \leq  q_i \tau_i (1 + \log t)
+ \sqrt{4 q_i \tau_i (1 + \log t) \log(\beta) + 5 \log^2(\beta)}. \label{eq:conc4} \\
& \text{In the adversarial model,} \notag\\
& \max_{i \in \{1,\hdots,K\}} \sum_{t=\tau_0+1}^n g_{i,t} - \sum_{t=\tau_0+1}^n g_{I_t,t} \leq 5.15 \sqrt{(n-\tau_0) K \log(\beta)}. \label{eq:conc5}
\end{align}
We will now make a deterministic reasoning on the event that the above inequalities are indeed true.

\subsection{Analysis in the stochastic model} \label{sec:proof2}

First note that by equations \eqref{eq:conc1} and \eqref{eq:conc3}, test \eqref{eq:test1} is never satisfied.

Let $i^* \in \argmax_i \mu_i$. Remark that by equation \eqref{eq:conc1}, test \eqref{eq:test2} is never satisfied for $i^*$, since if $i, i^* \in A_t$ then
$$\tH - \tilde{H}_{i^*,t} \leq - \Delta_i + 2 \sqrt{\frac{4 K \log(\beta)}{t} +
5 \left( \frac{K \log(\beta)}{t} \right)^2}.$$
Thus we have $i^* \in A_t$, $\forall t$.
Moreover if $i \not\in A_t$, then it means
that $\tau_i \leq t$ and test \eqref{eq:test2} was satisfied at time step $\tau_i$ (and not satisfied at time $\tau_i - 1$). Thus, using \eqref{eq:conc1}, we see that
if $i \not\in A_t$ then it implies:
$$\Delta_i + 2 \sqrt{\frac{4 K \log(\beta)}{\tau_i} +
5 \left( \frac{K \log(\beta)}{\tau_i} \right)^2} >
6 \sqrt{\frac{4 K \log(\beta)}{\tau_i} +
5 \left( \frac{K \log(\beta)}{\tau_i} \right)^2},$$
and (since $i^* \in A_t$)
\begin{equation}
 \Delta_i - 2 \sqrt{\frac{4 K \log(\beta)}{\tau_i-1} +
5 \left( \frac{K \log(\beta)}{\tau_i-1} \right)^2} \leq
6 \sqrt{\frac{4 K \log(\beta)}{\tau_i-1} +
5 \left( \frac{K \log(\beta)}{\tau_i-1} \right)^2} . \label{eq:LBtaui}
\end{equation}
Thus test \eqref{eq:test3} is never satisfied since:
$$
\max_{j \in A_t} \tilde{H}_{j,t} - \tH
\leq \Delta_i + 2 \sqrt{\frac{4 K \log(\beta)}{\tau_i} +
5 \left( \frac{K \log(\beta)}{\tau_i} \right)^2}
\leq 10 \sqrt{\frac{4 K \log(\beta)}{\tau_i-1} +
5 \left( \frac{K \log(\beta)}{\tau_i-1} \right)^2}.
$$
Moreover \eqref{eq:test4} is also never satisfied, indeed since $i^* \in A_t$ we have:
$$
\max_{j \in A_t} \tilde{H}_{j,t} - \tH
\geq \Delta_i - 2 \sqrt{\frac{4 K \log(\beta)}{\tau_i} +
5 \left( \frac{K \log(\beta)}{\tau_i} \right)^2}
>  2 \sqrt{\frac{4 K \log(\beta)}{\tau_i} +
5 \left( \frac{K \log(\beta)}{\tau_i} \right)^2}.
$$

In conclusion we proved that Exp3 is never started in the stochastic model, that is $\tau_0 = n$. Thus, using \eqref{eq:conc4}, we obtain:
\begin{eqnarray*}
\oR_n & = & \sum_{i=1}^K \Delta_i T_i(n) \\
& \leq & \sum_{i=1}^K \Delta_i
\left( q_i \tau_i (1 + \log n)
+ \sqrt{4 q_i \tau_i (1 + \log n) \log(\beta) + 5 \log^2(\beta)}\right).
\end{eqnarray*}
Now remark that for any arm $i$ with $\Delta_i>0$, one can see that \eqref{eq:LBtaui} implies:
$$\tau_i \leq 259 \frac{K \log(\beta)}{\Delta_i^2} + 1 \leq 260 \frac{K \log(\beta)}{\Delta_i^2}.$$
Indeed if $\tau_i > 259 \frac{K \log(\beta)}{\Delta_i^2} + 1$, then
$$8 \sqrt{\frac{4 K \log(\beta)}{\tau_i - 1} + 5 \left(\frac{K \log(\beta)}{\tau_i - 1}\right)^2} < 8 \sqrt{\frac{4 \Delta_i^2}{259} + \frac{5 \Delta_i^4}
{259}} < \Delta_i,$$
which contradicts \eqref{eq:LBtaui}.

The proof is concluded with straightforward computations and by showing that
\begin{equation} \label{eq:qi}
 \sum_{i=1}^K q_i \leq 1+\log K .
\end{equation}
Denote by $\tau_{(1)} \leq \hdots \leq \tau_{(K)}$ the ordered random variables $\tau_1,\hdots, \tau_K$. Then we clearly have $q_{(i)} \leq \frac{1}{K-i+1}$,
which proves \eqref{eq:qi}.

\subsection{Analysis in the adversarial model} \label{sec:proof3}

Let $i^* \in \argmax_{1 \leq i \leq K} G_{i,\tau_0-1}$. First we show that $i^* \in A_{\tau_0-1}$. Let $I^* \in \argmax_{i \in A_{\tau_0-1}} G_{i,\tau_0-1}$ and
$i \not\in A_{\tau_0-1}$,
then we have, by $\tau_i \leq \tau_0 - 1$, \eqref{eq:conc2} and since \eqref{eq:test4} is not satisfied at time $\tau_0-1$:
\begin{align*}
& G_{I^*,\tau_0-1} - G_{i,\tau_0-1} \\
& = G_{I^*,\tau_0-1} - \tilde{G}_{I^*,\tau_0-1}  + \tilde{G}_{I^*,\tau_0-1} - \tilde{G}_{i,\tau_0-1}  + \tilde{G}_{i,\tau_0-1} -
G_{i,\tau_0-1} \\
& > - \sqrt{4 \left(\frac{K \tau_i}{(\tau_0-1)^2} + \frac{\tau_0-1 - \tau_i}{q_i \tau_i (\tau_0-1)}\right) \log(\beta) +
5 \left( \frac{K \log(\beta)}{\tau_i} \right)^2} - \sqrt{\frac{4 K \log(\beta)}{\tau_0 - 1} +
5 \left( \frac{K \log(\beta)}{\tau_0 - 1} \right)^2} \\
& \qquad +
2 \sqrt{\frac{4 K \log(\beta)}{\tau_i} +
5 \left( \frac{K \log(\beta)}{\tau_i} \right)^2} \\
& \geq - \sqrt{4 \left(\frac{K \tau_i}{(\tau_0-1)^2} + \frac{\tau_0-1 - \tau_i}{q_i \tau_i (\tau_0-1)}\right) \log(\beta) +
5 \left( \frac{K \log(\beta)}{\tau_i} \right)^2} + \sqrt{\frac{4 K \log(\beta)}{\tau_i} +
5 \left( \frac{K \log(\beta)}{\tau_i} \right)^2},
\end{align*}
where the last inequality follows from $q_i \geq 1/K$ and
$$\frac{\tau_i}{(\tau_0 - 1)^2} + \frac{\tau_0 - 1 - \tau_i}{\tau_i (\tau_0 - 1)} \leq \frac{1}{\tau_i} .$$
This proves $i^* \in A_{\tau_0-1}$. Thus we get, using the fact that \eqref{eq:test1} and \eqref{eq:test3} are not satisfied at time $\tau_0-1$,
as well as \eqref{eq:conc2}, and the fact that \eqref{eq:test2} is not satisfied for active arms at time $\tau_0 - 1$,
\begin{eqnarray*}
R_{\tau_0-1}
& = & G_{i^*,\tau_0-1} - \sum_{i=1}^K \widehat{G}_{i,\tau_0-1} \\
& = & \sum_{i=1}^K T_i(\tau_0-1) \left(H_{i^*, \tau_0-1} - \widehat{H}_{i,\tau_0-1} \right) \\
& = & \sum_{i=1}^K T_i({{\tau_0-1}}) \left(H_{i^*,{{\tau_0-1}}}- \tilde{H}_{i^*,{{\tau_0-1}}}
+ \tilde{H}_{i^*,{{\tau_0-1}}} - \tilde{H}_{i,{{\tau_0-1}}}
+ \tilde{H}_{i,{{\tau_0-1}}} - \widehat{H}_{i,{{\tau_0-1}}} \right) \\
& \leq & \sum_{i=1}^K T_i(\tau_0-1) \left(12 \sqrt{\frac{4 K \log(\beta)}{\tau_i-1} +
5 \left( \frac{K \log(\beta)}{\tau_i-1} \right)^2}
+ \sqrt{\frac{2 \log(\beta)}{T_i(\tau_0-1)}} \right).
\end{eqnarray*}
Then, using \eqref{eq:conc4} and \eqref{eq:conc5} we get, thanks to $\tau_i \geq 2$,
\begin{eqnarray*}
R_n & \leq & 1 + 6.6 \sqrt{n K \log(\beta)} + 12 \sum_{i=1}^K q_i (1+\log n) \sqrt{16 K \tau_i \log(\beta) + 20 (K \log(\beta))^2} \\
& + &  12 \sum_{i=1}^K
\sqrt{\left(4 q_i \tau_i (1 + \log n) \log(\beta) + 5 \log^2(\beta)\right) \left(\frac{4 K \log(\beta)}{\tau_i-1} +
5 \left( \frac{K \log(\beta)}{\tau_i-1} \right)^2\right)} \\
& \leq & 60 (1+\log K) (1+ \log n)\sqrt{n K \log(\beta) + K^2 \log^2(\beta)} + 200 K^2 \log^2(\beta),
\end{eqnarray*}
where the last inequality follows from \eqref{eq:qi} and straightforward computations.

\xhdr{Acknowledgements.} We thank Peter Auer for insightful discussions.

\newpage
\appendix
\section{Concentration inequalities}
\label{app:chernoff}

Recall that the analysis in Section~\ref{sec:2-arms} relies on Chernoff Bounds as stated in Theorem~\ref{thm:our-chernoff}. Let us derive Theorem~\ref{thm:our-chernoff} from a version of Chernoff Bounds that can be found in the literature.

\begin{theorem}[Chernoff Bounds: Theorem 2.3 in~\cite{McDiarmid-concentration}]
\label{thm:chernoff-McDiarmid}
Consider $n$ i.i.d. random variables $X_1 \ldots X_n$ on $[0,1]$. Let
    $X=\tfrac{1}{n}\,\sum_{t=1}^n X_t$
be their average, and let $\mu = \E[X]$. Then for any $\eps>0$ the following two properties hold:
\begin{itemize}
\item[(a)] $\Pr[ X \geq (1+\eps) \mu ] < \exp\left( -\frac{\eps^2\mu}{2(1+\eps/3)} \right)
    < \begin{cases}
        e^{-\eps^2\mu/3}, & \eps\leq 1 \\
        e^{-\eps\mu/3}, & \text{otherwise}.
    \end{cases} $
\item[(b)] $\Pr[ X \leq (1-\eps) \mu ] < e^{-\eps^2\mu/2}$.
\end{itemize}
\end{theorem}

\begin{corollary}
In the setting of Theorem~\ref{thm:chernoff-McDiarmid}, for any $\beta>0$ we have:
\begin{align}\label{eq:my-chernoff-pf}
\Pr\left[\, |X - \mu| > \beta\,\max(\beta,\sqrt{\mu}) \,\right] < 2\,e^{-\beta^2/3}.
\end{align}
We obtain Theorem~\ref{thm:our-chernoff} by taking $\beta = \sqrt{C}$, noting that
    $ \beta\,\max(\beta,\sqrt{\mu}) \leq C\max(1,\sqrt{\mu})$
for $C>1$.
\end{corollary}

\begin{proof}
Fix $\beta>0$ and consider two cases: $\mu\geq \beta^2$ and $\mu< \beta^2$.

If $\mu\geq \beta^2$ then we can take
    $\eps = \beta/\sqrt{\mu}\leq 1$
in Theorem~\ref{thm:chernoff-McDiarmid}(ab) and obtain
$$ \Pr[ |X-\mu| \geq \beta \sqrt{\mu} ]
    = \Pr[ |X-\mu| \geq \eps \mu ] < 2\,e^{-\eps^2\mu/3} = 2\,e^{-\beta^2/3} .$$

Now assume $\mu<\beta^2$. We can still take $\eps = \beta/\sqrt{\mu}$ in Theorem~\ref{thm:chernoff-McDiarmid}(b) to obtain
$$ \Pr[ X-\mu \leq - \beta^2 ]
   \leq \Pr[ X-\mu \leq - \beta\sqrt{\mu} ] < e^{-\eps^2\mu/2} = e^{-\beta^2/2}.$$
Then let us take $\eps = \beta^2/\mu>1$ in Theorem~\ref{thm:chernoff-McDiarmid}(a) to obtain
$$ \Pr[ X-\mu \geq \beta^2 ]
    = \Pr[ X-\mu \geq \eps \mu ] < e^{-\eps \mu/3} = e^{-\beta^2/3}.$$
It follows that
    $\Pr[\, |X-\mu|\geq \beta^2 \,] < 2\,e^{-\beta^2/3}$,
completing the proof.
\end{proof}

\end{document}